\documentclass[final]{cvpr}

\usepackage{times}
\usepackage{epsfig}
\usepackage{graphicx}
\usepackage{amsmath}
\usepackage{amssymb}
\usepackage{caption}
\usepackage{subcaption}
\usepackage{multirow}
\usepackage{booktabs}
\usepackage{bbm}
\usepackage{enumitem}
\usepackage{placeins}
\usepackage{nopageno}
\usepackage[pagebackref=true,breaklinks=true,colorlinks,bookmarks=false]{hyperref}
\newcommand\blfootnote[1]{%
  \begingroup
  \renewcommand\thefootnote{}\footnote{#1}%
  \addtocounter{footnote}{-1}%
  \endgroup
}

\def\cvprPaperID{7055} 
\def\confYear{CVPR 2021}

\begin{document}

\title{Interpretable Social Anchors for Human Trajectory Forecasting in Crowds}

\author{Parth Kothari, Brian Sifringer, Alexandre Alahi\\
EPFL VITA lab\\
CH-1015 Lausanne\\
{\tt\small parth.kothari@epfl.ch}

}

\maketitle

\begin{abstract}
Human trajectory forecasting in crowds, at its core, is a sequence prediction problem with specific challenges of capturing inter-sequence dependencies (social interactions) and consequently predicting socially-compliant multimodal distributions. In recent years, neural network-based methods have been shown to outperform hand-crafted methods on distance-based metrics. However, these data-driven methods still suffer from one crucial limitation: lack of interpretability. To overcome this limitation, we leverage the power of discrete choice models to learn interpretable rule-based intents, and subsequently utilise the expressibility of neural networks to model scene-specific residual. Extensive experimentation on the interaction-centric benchmark TrajNet++ demonstrates the effectiveness of our proposed architecture to explain its predictions without compromising the accuracy.\and
\end{abstract}

\section{Introduction}

\blfootnote{To appear in Computer Vision and Pattern Recognition (CVPR) 2021} 
Humans naturally navigate through crowds by following the unspoken rules of social motion such as avoiding collisions or yielding right-of-way. Forecasting human motion in public places is a challenging, yet crucial task for the success of many applications like deployment of autonomous navigation systems \cite{WaymoSafety, UberSafety, Chen2019CrowdRobotIC}, infrastructure design \cite{Jiang1999SimPedSP, Lerner2007CrowdsBE} and evacuation analysis \cite{Helbing2002SimulationOP, Zheng2009ModelingCE}. Therefore, in the last few decades, developing models that can understand human social interactions and forecast future trajectories has been an active and challenging area of research. 

Early works designed hand-crafted methods based upon domain knowledge to forecast human trajectories, either with physics-based models such as Social Forces \cite{SocialForce}, or with pattern-based models such as discrete choice modelling (DCM) \cite{antonini2006discrete, GUO2008580, asano2010microscopic}. These models, based on domain knowledge, were successful in showcasing crowd phenomena like collision avoidance and leader-follower type behavior. Moreover, the hand-designed nature of these models rendered their predictions to be interpretable. However, human motion in crowds is much more complex and due to its long-term nature, these first-order methods suffer from predicting inaccurate trajectories. 

\begin{figure}
\centering
\begin{subfigure}[h]{0.45\textwidth}
    \includegraphics[width=\textwidth]{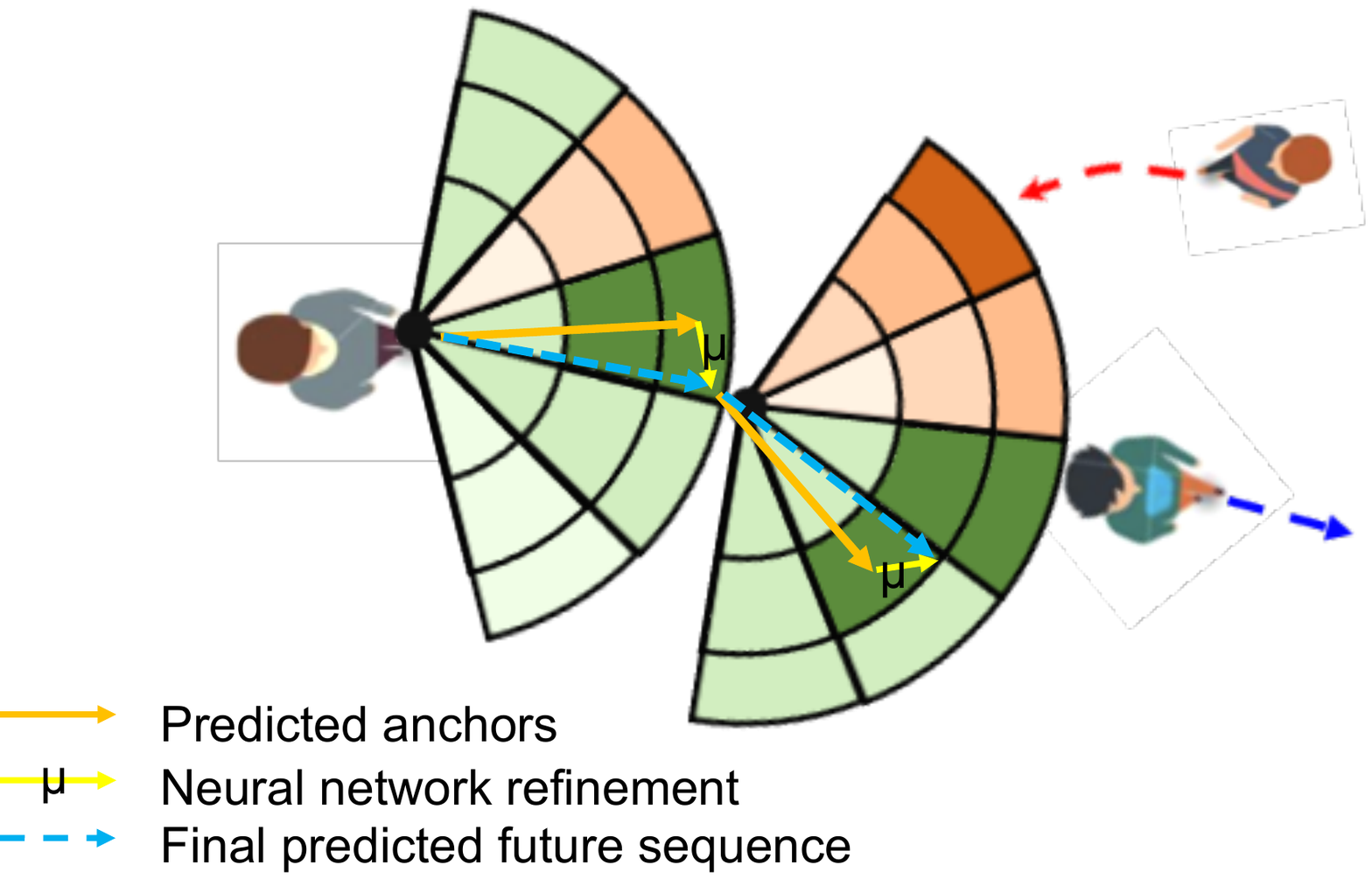}
\end{subfigure}
\setlength{\belowcaptionskip}{-12pt}
\caption{While navigating in crowds, humans display various social phenomena like collision avoidance (from red trajectory) and leader follower (towards blue trajectory). We present a model that not only outputs accurate future trajectories but also provides a high-level rationale behind its predictions, owing to the interpretability of discrete choice models. (Un)favourable anchors shown in green (red).}
%
\label{fig:pull}
\end{figure}

Building on the success of recurrent neural network-based models in learning complex functions and long-term dependencies, Alahi \textit{et al.} \cite{Alahi2016SocialLH} proposed the first neural network (NN) based trajectory forecasting model, Social LSTM, which outperformed the hand-crafted methods on distance-based metrics. Due to the success of Social LSTM, neural networks have become the de-facto choice for designing human trajectory models \cite{ Gupta2018SocialGS, Zhang2019SRLSTMSR, Zhu2019StarNetPT, Ivanovic2018TheTP, Giuliari2020TransformerNF}. However, current NN-based trajectory forecasting models suffer from a significant limitation: lack of interpretability regarding the model's decision-making process. 

In this work, we are interested in combining the forces of the two paradigms of human trajectory forecasting (see Fig.~\ref{fig:pull}): the interpretability of the trajectories predicted by hand-crafted models, in particular discrete choice models \cite{antonini2006discrete,ROBIN200936}, and the high accuracy of the neural network-based predictions. With this objective, we propose a model that outputs a probability distribution over a discrete set of possible future intents. This set is designed as a function of the pedestrian's speed and direction of movement. Our model learns the probability distribution over these intents with the help of a choice model architecture, owing to its ability to output interpretable decisions. To this end, we resort to a novel hybrid and interpretable framework in DCM \cite{sifringer2020enhancing}, where knowledge-based hand-crafted functions can be augmented with neural network representations, without compromising the interpretability.

Our architecture augments each predicted high-level intent with a scene-specific residual term generated by a neural network. The advantage of this is two-fold: first, the residual allows to expand the output space of the model from a discrete distribution to a continuous one. Secondly, it helps to incorporate the complex social interactions as well as the long-term dependencies that the first-order hand-crafted models fail to capture, leading to an increase in prediction accuracy. Overall, we can view our architecture as disentangling high-level coarse intents and lower-level scene-specific nuances of human motion. 



We demonstrate the efficacy of our proposed architecture on TrajNet++ \cite{Kothari2020HumanTF}, an interaction-centric human trajectory forecasting benchmark comprising of well-sampled real-world trajectories that undergo various social phenomena. Through extensive experimentation, we demonstrate that our method performs at par with competitive baselines on both real-world and synthetic datasets, while at the same time providing a rationale behind high-level decisions, an essential component required for safety-critical applications like autonomous systems.


\section{Related Work}



\subsection{Social Interactions}

Current human trajectory forecasting research can be categorized into learning human-human (social) interactions and human-space (physical) interactions. In this work, we focus on the task of designing models that aim at understanding social interactions in crowds. The human social interactions are usually modelled either using knowledge-based models or using neural networks. 


\textbf{Knowledge-based Models}: With a specific focus on pedestrian path forecasting problem, Helbing and Molnar \cite{SocialForce} presented a force-based motion model with attractive forces (towards the goal and one's own group) and repulsive forces (away from obstacles), called Social Force model. Burstedde \textit{et al.} \cite{Burstedde2001SimulationOP} utilize the cellular automaton model to predict pedestrian motion by dividing the environment into uniform grids and assigning transition preference matrices to the pedestrians. Similarly, discrete choice modelling utilizes a grid for selecting the next action, but relative to each individual \cite{antonini2006discrete, ROBIN200936, ondvrej2010synthetic}. The high interpretability and design flexibility of DCM allowed its application to many topics such as pedestrian flows \cite{liu2014agent}, walking in groups \cite{moussaid2010walking, yamaguchi2011you}, collision avoidance \cite{asano2010microscopic, liu2015modeling}, and critical or emergency situations \cite{GUO2008580,moussaid2011simple, xie_cooperative_2016}. Human social interactions have also been modelled using other knowledge-based perspectives \cite{Berg2008ReciprocalVO, Robicquet2016LearningSE, Alahi2014SociallyAwareLC}. While the hand-crafted functions of these methods lead to interpretable outputs, they are often too simple to capture the complexity of human interactions. 

\textbf{Neural Network-based Models: } In the past few years, methods based on neural networks (NNs) that infer social interactions in a data-driven fashion have been shown to outperform the knowledge-based works on distance-based metrics.  Social LSTM \cite{Alahi2016SocialLH} introduced a novel social pooling layer to capture social interactions of nearby pedestrians. Various other interaction-capturing NN modules have been proposed in literature \cite{Pfeiffer2017ADM, Shi2019PedestrianTP, Bisagno2018GroupLG, Gupta2018SocialGS, Zhang2019SRLSTMSR, Zhu2019StarNetPT, Ivanovic2018TheTP, Liang2019PeekingIT, Tordeux2019PredictionOP}. To provide different weights to neighbours that affect the trajectory of the pedestrian of interest, multiple works \cite{Xu2018EncodingCI, Fernando2018SoftH, Li2020SocialWaGDATIT, Sadeghian2018SoPhieAA, Kosaraju2019SocialBiGATMT, Amirian2019SocialWL, Fernando2018GDGANGA, Haddad2019SituationAwarePT, Huang2019STGATMS, Mohamed2020SocialSTGCNNAS, Li2020EvolveGraphHM} propose to utilize attention mechanisms \cite{Vaswani2017AttentionIA, Bahdanau2014NeuralMT}. The attention weights are either learned or handcrafted based on domain knowledge (\textit{e.g.}, euclidean distance). However, these data-driven methods lack the ability to output predictions that can be explained, unlike their knowledge-based counterparts.


In this work, we combine the strengths of rule-based models to output high-level intents that are interpretable, and NN-based models to predict scene-specific residuals that take into account the long-term motion characteristics.

\begin{figure*}[h]
\centering
\begin{subfigure}[h]{\textwidth}
    \includegraphics[width=\textwidth]{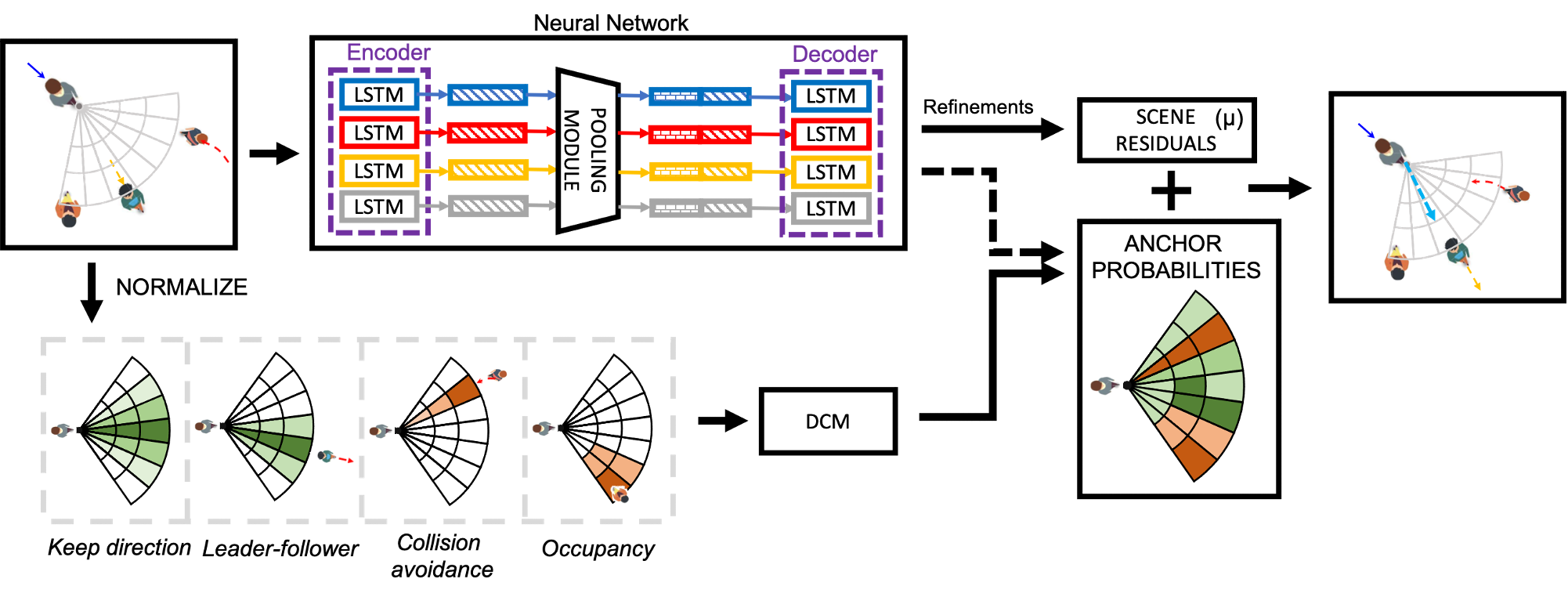}
\end{subfigure}

\caption{At each time-step, the output space of each pedestrian is discretized into a set of possible future intents, normalized with respect to the pedestrian's speed and direction, forming a radial grid. Discrete choice modelling (DCM) is used to predict the next step probability distribution (green high, red low) in an interpretable manner by accumulating the \textit{keep direction, leader-follower, collision avoidance} and \textit{occupancy} rules. A neural network refines the predicted anchor distribution with scene-specific residuals that account for the subtle interactions that the DCM rules fail to model. The neural network also provides the past motion embedding and interactions embedding which can be added to the hand-crafted DCM functions to better handle complex social interactions and long term dependencies while choosing the future intents.}
\label{fig:synth_traj}
\end{figure*}

\subsection{Multimodality}
 
Training neural networks based on minimization of $L_2$ loss leads to the model outputting the mean of all the possible outcomes. One solution to ensure multimodal forecasting is to explicitly output multiple modes using the decoder architecture, for instance, using Mixture Density Networks \cite{Bishop1994MixtureDN}. However, this training technique suffers from numerical instabilities, often leading to mode collapse.

Another recently popular approach is based on generative modelling \cite{Lee2017DESIREDF, Ivanovic2018TheTP, Gupta2018SocialGS, Amirian2019SocialWL, Li2019WhichWA}. Generative models implicitly model the probability distribution of the future trajectories conditioned on the past scene, thereby naturally offering a possibility to output multiple samples. Lee \textit{et al.} \cite{Lee2017DESIREDF} propose a recurrent encoder-decoder architecture within conditional variational autoencoder (cVAE) framework. Ivanovic \textit{et al.} \cite{Ivanovic2018TheTP} propose to use Gaussian mixture model (GMM) on top of the recurrent decoder in cVAE framework. Several works \cite{Gupta2018SocialGS, Amirian2019SocialWL, Li2019WhichWA} utilize generative adversarial networks to model trajectory distributions. Gupta \textit{et al.} \cite{Gupta2018SocialGS} utilize Winner-takes-all (WTA) \cite{Rupprecht2016LearningIA} loss, in addition to adversarial loss, to encourage the network to produce diverse samples covering all modes. Amirian \textit{et al.}\cite{Amirian2019SocialWL} propose to use InfoGAN architecture to tackle mode collapse. 


In this work, we recast the problem of multimodality as learning a distribution over the agent's intents. We predict the distribution of these high-level intents by leveraging the interpretability of choice models. Therefore, unlike previous works, our model explicitly provides a rationale and a ranking for each future mode. 

\section{Method}


Humans have mastered the ability to negotiate complicated social interactions by anticipating the movements of surrounding pedestrians, leading to social concepts such as collision avoidance and leader-follower. Current NN-based architectures, despite displaying high accuracy, are unable to provide a rationale behind their accurate predictions. Our objective is to equip these models with the ability to provide a social concept-based reason behind their decisions. In this section, we describe our proposed architecture, that outputs a high-level intent and a scene-specific residual corresponding to each intent, followed by our DCM-based component that makes the intent interpretable.

\subsection{Problem Definition}
For a particular scene, we receive as input the trajectories of all people in a scene as $\mathbf{X} = [X_1, X_2, ... X_n]$, where $n$ is the number of people in the scene. The trajectory of a person $i$, is defined as $X_i = (x_i^t,y_i^t)$, from time $t=1,2...t_{obs}$ and the future ground-truth trajectory is defined as $Y_i = (x_i^t,y_i^t)$ from time $t=t_{obs}+1,...t_{pred}$. The goal is to accurately and simultaneously forecast the future trajectories of all people $\mathbf{\hat{Y}}=[\hat{Y}_1,\hat{Y}_2...\hat{Y}_n]$, where $\hat{Y}_i$ is used to denote the predicted trajectory of person $i$. The velocity of a pedestrian $i$ at time-step $t$ is denoted by $\mathbf{v}^{t}_{i}$.

\subsection{Discrete Choice Models}
The theory of discrete choice models (DCMs) is built on a strong mathematical framework, allowing high interpretability regarding the decision making process \cite{mcfadden1973conditional}. DCMs have often been applied in fields of economy \cite{aguirregabiria2010dynamic}, health \cite{ryan2007using} and transportation \cite{bhat2007flexible}, where interpretation of parameters that capture human behavior is of utmost importance. These models are used to predict, for each person $i$, their choice among an available set of $K$ options. In the most common DCM-based approach, called Random Utility Maximization (RUM), \cite{manski1977structure}, each option has an associated hand designed function $u_k$, called utility, and each person is assumed to select the option for which their utility is maximized.

The inputs $\Tilde{\mathbf{x}}$ to these utility functions are designed via expert knowledge of the given problem, and are then assigned a vector of learnable weights $\beta$. These weights are regressed on all available options with respect to the observations, and reflect the impact of each component in the utility function. It is the study of these weights and corresponding input values that lead to the high interpretability of discrete choice methods at the individual and population level. Formally, the utility for option $k$ is calculated as:
\begin{equation}
    U_{k}(\beta, \Tilde{\mathbf{x}}) = u_k(\beta,\Tilde{\mathbf{x}}) + \varepsilon_{k}
\end{equation}
where $\varepsilon_{k}$ is the random term. Varying assumptions on the distribution of this random term leads to different types of DCM models \cite{williams1977formation, mcfadden1978modeling}. 

While many works incorporate data-driven methods into the DCM framework \cite{bentz2000neural, hruschka2004empirical, wong2020bi}, only recently have models been proposed that keep the knowledge-based functions and the parameters interpretable after adding the neural network \cite{sifringer2020enhancing, han2020neural}. In this paper, we utilize the Learning Multinomial Logit (L-MNL) \cite{sifringer2020enhancing}, as our base DCM model. 


\subsection{Model Architecture}\label{scene_res}
As shown in Figure~\ref{fig:synth_traj}, at each time-step, our model outputs a distribution over a discretized set of $K$ future intents, which we term  \textit{social anchors}, denoted by $\mathcal{A} = \{a_k\}_{k=1}^{K}$, as well as scene-specific refinements for each intent. The size of the set is defined by the number of speed levels $N_s$, and direction changes $N_d$ such that $K=N_d\times N_s$.  As we will see in Section \ref{sec:anchor_select}, we choose to utilize a DCM to output the distribution over these anchors because of the its ability to explain its decisions. Next, we utilize the high expressibility of neural networks to provide a refinement in the output space with respect to each anchor in $\mathcal{A}$. We call these refinements \textit{scene residuals}. These scene-specific residuals allow us to project the coarse and discretized problem back into the continuous domain. Note that the set $\mathcal{A}$ is chosen to be rich enough to provide a desired level of coverage in the output space, so that the magnitudes of the scene-specific residuals are minimal.

\textbf{Scene Residuals: }
We now describe our neural network architecture that is used to output scene-specific residuals corresponding to each anchor. These residuals are used to model the long-term motion dependencies as well as the complex and often subtle social interactions that cannot be described using first-order hand-crafted rules. 
We first embed the velocity $\mathbf{v}^{t}_{i}$ of pedestrian $i$ at time $t$ using a single layer MLP to get a fixed length embedding vector $\mathbf{e}^{t}_{i}$ given as:
\begin{equation}
    e^t_i = \phi_{emb}(\textbf{v}^t_i; W_{emb}),
\end{equation}
where $\phi_{emb}$ is the embedding function, $W_{emb}$ are the weights to be learned.


Next, we utilize the directional pooling module proposed in \cite{Kothari2020HumanTF} to model the social interactions and obtain the interaction vector $p^t_{i}$. We then concatenate the input embedding with the interaction embedding and provide the concatenated vector as input to the LSTM module, obtaining the following recurrence:
\begin{equation} \label{eq:LSTM_main}
    h^t_{i} = LSTM(h^{t-1}_i, [e^t_{i}; p^{t}_{i}]; W_{\mathrm{encoder}}),
\end{equation}
where $h^t_{i}$ denotes the hidden state of pedestrian $i$ at time $t$, $W_{\mathrm{encoder}}$ are the weights to be learned. The weights are shared between all pedestrians in the scene.

The hidden-state at time-step $t$ of pedestrian $i$ is then used to predict the residuals corresponding to each anchor at time-step $t + 1$. Similar to \cite{Alahi2016SocialLH}, we characterize the residual corresponding to the $k^{th}$ anchor as a bivariate Gaussian distribution parameterized by the mean  $\mu_k^{t+1} = (\mu_x,\mu_y)_k^{t+1}$, standard deviation $\sigma_k^{t+1} = (\sigma_x,\sigma_y)_k^{t+1}$ and correlation coefficient $\rho_k^{t+1}$:

\begin{equation}
    [\mu_k^{t}, \sigma_k^{t}, \rho_k^{t}] = \phi_{dec}(h_i^{t-1}, W_{\mathrm{decoder}}),
\end{equation}
where $\phi_{dec}$ is modelled using an MLP and $W_{decoder}$ is learned.

\subsection{Anchor Selection}\label{sec:anchor_select}
The pedestrian's intent for the next time-step is discretized as a set of $K$ future intentions $\mathcal{A} = \{a_k\}_{k=1}^{K}$. The selection of an anchor from the choice set $\mathcal{A}$ is posed as a discrete choice modelling task. This is made possible by normalizing the anchor set with respect to both a person's speed and direction. We describe the role of normalization to integrate the DCM structure in Sec.~\ref{sec:dir_norm}.

While many different rules and behaviors for human motion have been described in DCM literature, we follow the formulation described in \cite{antonini2006discrete, ROBIN200936}, which is well adapted to our problem setting. Functions modelling human motion phenomenon which we consider for anchor selection in this work are:


\begin{enumerate}[itemsep=0pt,parsep=0pt,topsep=0pt]
\item \textbf{\textit{avoid occupancy}}: directions containing neighbours in the vicinity are less desirable, scaled by the inverse-distance to the considered anchor.
    
\item \textbf{\textit{keep direction}}: pedestrians tend to maintain the same direction of motion.


\item \textbf{\textit{leader-follower}}: pedestrians have a tendency to follow the tracks of people heading in the same direction, identified as `leaders'. The relative speed of the leader with respect to the follower entices the follower to slow down or accelerate.
    
\item \textbf{\textit{collision avoidance}}: when a neighbour pedestrian's trajectory is head-on towards an anchor, this anchor becomes less desirable due to the chance of a collision. 
    
    
\end{enumerate}
An illustration of the effects of the above chosen functions on the final anchor selection is shown in Figure \ref{fig:synth_traj}. Given the chosen functions, the associated utility $u_k$ for anchor $k$ is written as:
\begin{align}
     u_k(\mathbf{X}) =& \underbrace{\beta_{dir} dir_k}_{\textit{keep direction}} + \underbrace{\beta_{occ} occ_k}_{\textit{avoid occupancy}} + \underbrace{\beta_C col_k}_{\textit{collision avoidance}} \nonumber \\
    & \underbrace{+ \beta_{acc} L_{k,acc} + \beta_{dec}  L_{k,dec}}_{\textit{leader-follower}} \; , \label{eq:utility}
\end{align}
where  $\beta_j$ are the learnable weights of the corresponding functions. The exact mathematical formulations of the above functions are detailed in \cite{ROBIN200936, antonini2006discrete}.  Each person is assumed to select the anchor $a_k$ for which the corresponding utility $u_k$ is maximum. 

We would like to point that the performance of the underlying DCM is determined by the hand-crafted functions of human motion that it models. The DCM framework provides the flexibility to integrate any other knowledge-driven function extensively tested in past literature. 

Although the knowledge-based functions offer stable and interpretable results, they are unable to capture the heterogeneity of trajectory decisions in more complex situations. The future intent of a pedestrian is also dependent on long-term dependencies and subtle social interactions that these first-order hand-designed functions are unable to capture. The inclusion of NN-based terms helps to alleviate this issue.


Recently proposed L-MNL \cite{sifringer2020enhancing} architecture allows having both NN-based and knowledge-based terms in the utility while maintaining interpretability. We therefore utilize this framework and add an encoded map of past observations $h(\mathbf{X})$ to adjust for the lack of long term dependencies of knowledge-based equations. Similarly, we also add an encoded map of social interactions $p(\mathbf{X})$ with information from all the neighbours to help model complex interactions, otherwise not captured by hand-designed functions.


In summary, we formulate the anchor selection probabilities as follows:
\begin{align}
     \pi(a_k |\mathbf{X}) = \frac{e^{{s_k(\mathbf{X})}}}{\sum\limits_{j\in K}e^{{s_j(\mathbf{X})}}},
\end{align}
where 
\begin{equation}
         s_k(\mathbf{X})= u_k(\mathbf{X}) + h_{k}(\mathbf{X}) + p_{k}(\mathbf{X}).
\end{equation}
$s_k(\mathbf{X})$ represents the anchor function containing the NN encoded terms, $h_{k}(\mathbf{X})$ and $p_{k}(\mathbf{X})$, as well as the hand-designed term $u_k(\mathbf{X})$ (Eq.~\ref{eq:utility}), following the L-MNL framework. Note that we use DCM assumptions from L-MNL for measuring the anchor probabilities, rather than those of the cross-nested logit model in \cite{ROBIN200936}.

\textbf{Training:} All the parameters of our model are learned with the objective of minimizing the negative log-likelihood (NLL) loss:
\begin{equation}
   \small \log p(\textbf{y} | \mathbf{X}) = \sum_{t} \log \left( \sum_{k} \pi(a_k | \mathbf{X} ) \; \mathcal{N}(y^t |\nu^t_k, \; \Sigma^t_k) \right),
\end{equation}
with 
\begin{equation}
    {\nu^t_k}  = y^{t-1} + a_k + \mu^t_k, 
\end{equation}
and where $\mu^t_k$ and $\Sigma^t_k$ are the scene-specific residuals (described in Sec.~\ref{scene_res}), $a_k$ are the coordinates of anchor $k$ and $y^{t-1}$ is the last position preceding the prediction.

As mentioned earlier, given an anchor set $\mathcal{A}$ such that it sufficiently covers the output space, the magnitude of NN-based refinements are minimal. During training, we choose to penalize the anchor that is closest to the ground-truth velocity at each time-step. 


Therefore, we optimize the following function during training:

\begin{equation}
\begin{split}
l(\theta) = \sum_{t} \sum_{k}  \big[ \mathbbm{1}(k^t & = \hat{k}^{t}_{m}) \big( \log \pi(a_k |\mathbf{X})\\ & +  \log \mathcal{N}(y^t | \nu^t_k,  \Sigma^t_k) \big) \big],
\end{split}
\end{equation}


where $\mathbbm{1}(\cdot)$ is the indicator function, and $\hat{k}^{t}_{m}$ is the index of the anchor most closely matching the ground-truth trajectory $\mathbf{Y}$ at time $t$, measured as $l_2$-norm distance in state-sequence space. 

\textbf{Testing:} During test time, till time-step $t_{obs}$, we provide the ground truth position of all the pedestrians as input to the forecasting model. From time $t_{obs + 1}$ to $t_{pred}$, we use the predicted position (derived from the most-probable intent combined with the corresponding residual) of each pedestrian as input to the forecasting model and predict the future trajectories of all the pedestrians. 

\subsection{Implementation Details}

The velocity of each pedestrian is embedded into a 64-dimensional vector. The dimension of the interaction vector is fixed to 256. We utilize directional pooling \cite{Kothari2020HumanTF} as the interaction module in all the methods for a fair comparison, with a grid of size $16 \times 16$ with a resolution of $0.6$ meters. We perform interaction encoding at every time-step. The dimension of the hidden state of both the encoder LSTM and decoder LSTM is 256. Each pedestrian has their encoder and decoder. The batch size is fixed to 8. We train using ADAM optimizer \cite{Kingma2015AdamAM} with a learning rate of 1e-3 for 25 epochs. For the DCM-based anchor selection, all exponential parameters of the chosen hand-designed functions are set to the estimated values in \cite{antonini2006discrete, ROBIN200936}. For synthetic data, we embed the goals in a 64-dimensional vector.

\section{Experiments}

In this section, we highlight the ability of our proposed method to output socially-compliant interpretable predictions. We evaluate our method on the recently released interaction-centric TrajNet++ dataset. TrajNet++ dataset consists of real-world pedestrian trajectories that are carefully sampled such that the pedestrians of interest undergo social interactions and no collisions occur in both the training and testing set. In total there are around 200k scenes in challenging crowded settings showcasing group behavior, people crossing each other, collision avoidance and groups forming and dispersing. 

\vspace{1em}

\textbf{Evaluation}: we consider the following metrics:
\begin{enumerate}[itemsep=1pt,parsep=0pt,topsep=1pt]
  \item \textbf{Average Displacement Error (ADE)}: the average $L_{2}$ distance between ground-truth and model prediction overall predicted time steps. 
  \item \textbf{Final Displacement Error (FDE)}: the distance between the final predicted destination and the ground-truth destination at the end of the prediction period.
   \item \textbf{Collision I - Prediction collision (Col-I)} \cite{Kothari2020HumanTF}: this metric calculates the percentage of collision between the pedestrian of interest and the neighbours in the \textit{predicted} scene. This metric indicates whether the predicted model trajectories collide, \textit{i.e.}, whether the model learns the notion of collision avoidance.
   \item \textbf{Top-3 ADE/FDE}: given 3 output predictions for an observed scene, this metric calculates the ADE/FDE of the prediction \textit{closest} to the ground-truth trajectory in terms of ADE.
\end{enumerate}

\vspace{1em}
\textbf{Baselines}: we compare against the following baselines:
\begin{enumerate}[itemsep=1pt,parsep=0pt,topsep=1pt]
\item \textbf{S-LSTM}: we compare to S-LSTM \cite{Alahi2016SocialLH} baseline that outputs a unimodal trajectory distribution. 
\item \textbf{Winner-Takes-All (WTA)}: this architecture was proposed in \cite{Rupprecht2016LearningIA} to encourage the network to output diverse trajectories.
\item \textbf{SGAN}: Social GAN \cite{Gupta2018SocialGS}, a popular generative model to tackle multimodal trajectory forecasting.
\item \textbf{CVAE}: the Conditional Variational Auto-Encoder architectures has been shown recently \cite{Ivanovic2018TheTP, Lee2017DESIREDF} to successfully predict multi-modal trajectories by learning a sampling model given past observations.
\item \textbf{MinK}: to demonstrate the need for a fixed set of anchors, we compare against this baseline that directly outputs the NN residuals without any prior anchors.
\item \textbf{SAnchor} [\textbf{Ours}]: our proposed method that utilizes 15 anchors (5 angle profiles and 3 speed profiles) to predict multimodal trajectory distribution.
\end{enumerate}

\vspace{1em}

Table~\ref{Synth_456_comparison} and Table~\ref{Synth_456_comparison1} illustrate the quantitative performance of our proposed anchor-based method on TrajNet++ synthetic and real-world dataset respectively. Our method offers the advantage of providing interpretable predictions (discussed next) without compromising the accuracy on distance-based metrics against competitive baselines.

\begin{table}[h]
\centering
\resizebox{0.45\textwidth}{!}{
\begin{tabular}{ |c|c|c|c| } 
 \hline
\textbf{Model} & \textbf{ADE / FDE} & \textbf{Col-I} & \textbf{Top-3 ADE / FDE}\\ 
 \hline
 S-LSTM \cite{Alahi2016SocialLH} & 0.25/0.50 & 1.2 & 0.25/0.50* \\
  \hline
 WTA \cite{Rupprecht2016LearningIA}  &  0.28/0.54 & 4.8	& 0.22/0.42 \\
 \hline
 SGAN \cite{Gupta2018SocialGS} &  0.27/0.54 & 5.1	& 0.22/0.43 \\
 \hline
 CVAE \cite{Lee2017DESIREDF} & 0.26/0.52 & 1.9 & 0.23/0.47 \\
 \hline
 MinK & 0.34/0.72 & 5.2 & 0.22/0.42 \\
 \hline
 SAnchor [Ours] & \textbf{0.22/0.45} & \textbf{0.4} & \textbf{0.19/0.38} \\
 \hline
\end{tabular}}
\setlength{\belowcaptionskip}{0pt}
\caption{Performance on TrajNet++ synthetic data. Errors reported are ADE / FDE in meters, Col I in \%. We observe the trajectories for 9 times-steps (3.6 secs) and perform prediction for the next 12 (4.8 secs) time-steps. *Unimodal}
\label{Synth_456_comparison}
\end{table}

\begin{table}[h]
\centering
\resizebox{0.45\textwidth}{!}{
\begin{tabular}{ |c|c|c|c| } 
 \hline
\textbf{Model} & \textbf{ADE / FDE} & \textbf{Col-I} & \textbf{Top-3 ADE / FDE}\\ 
 \hline
 S-LSTM \cite{Alahi2016SocialLH} & \textbf{0.57/1.24} & 5.5 & 0.57/1.24* \\
  \hline
 WTA \cite{Rupprecht2016LearningIA} & 0.65/1.46 & 5.1 & \textbf{0.49/1.05} \\
 \hline
 SGAN \cite{Gupta2018SocialGS} & 0.66/1.45 & 5.9 & 0.51/1.08 \\
 \hline
 CVAE \cite{Lee2017DESIREDF} & 0.60/1.28 & 5.7 & 0.55/1.20 \\
 \hline
 MinK & 0.68/1.48 & 8.4 & 0.59/1.25 \\
 \hline
 SAnchor & 0.62/1.32 & \textbf{4.2} & 0.58/1.24 \\
 \hline
\end{tabular} }
\setlength{\belowcaptionskip}{-5pt}
\caption{Performance on TrajNet++ real data. Errors reported are ADE / FDE in meters, Col I in \%.  We observe the trajectories for 9 times-steps (3.6 secs) and perform prediction for the next 12 (4.8 secs) time-steps. *Unimodal}
\label{Synth_456_comparison1}
\end{table}


\subsection{Interpretability of the Intents}
The advantage of incorporating a discrete choice framework for predicting a pedestrian's next intended position is its interpretability. Our proposed architecture allows us to compare the hand-designed features extensively studied in literature along with the data-driven features to identify the most relevant factors, at a given time-step, for the anchor selection.

\begin{figure*}[ht]
\centering
\begin{subfigure}[h]{\textwidth}
    \includegraphics[width=\textwidth]{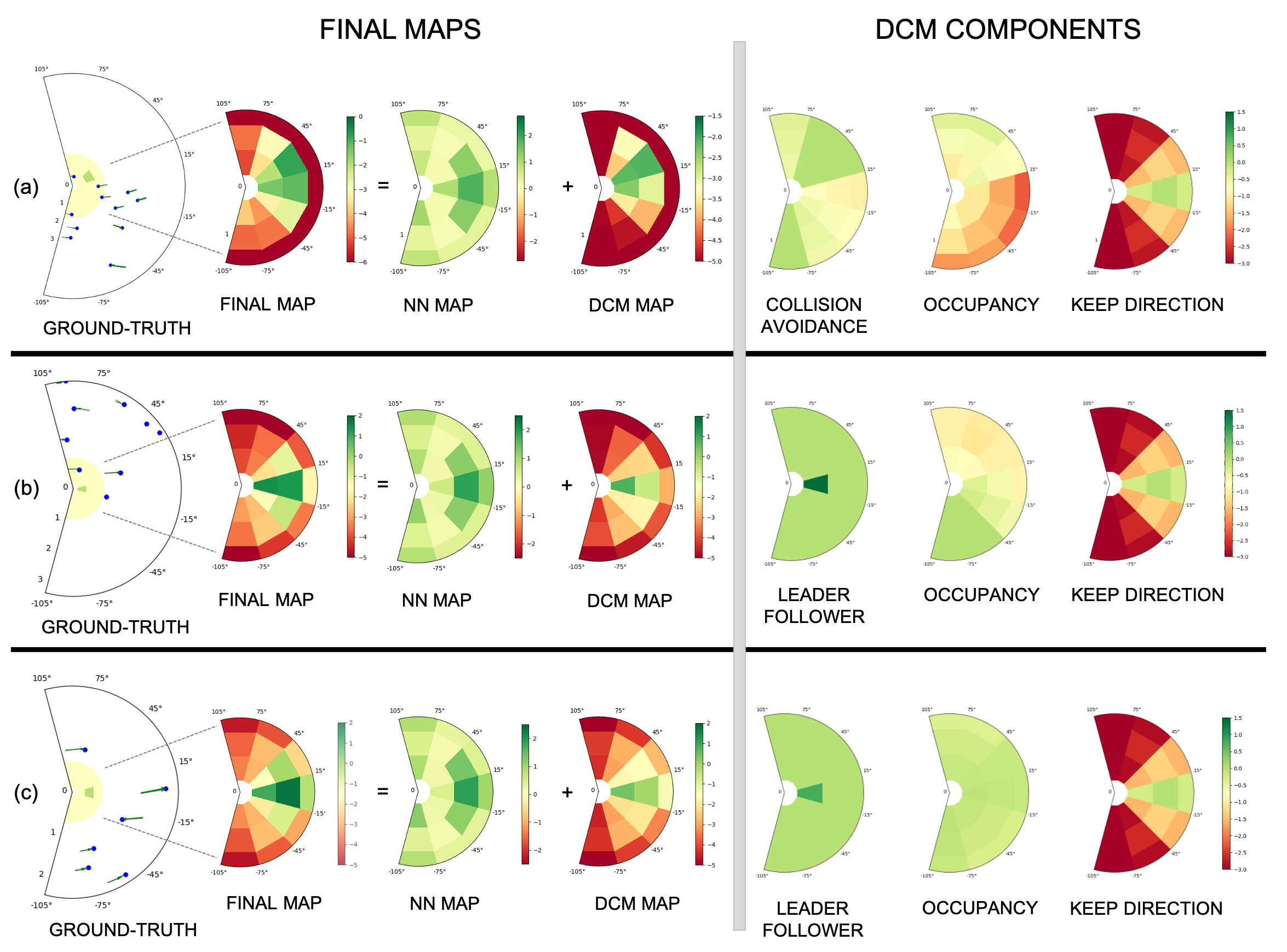}
\end{subfigure}
\caption{Qualitative illustration of the ability of our architecture to output high-level interpretable intents. The direction of motion of the pedestrian of interest is normalized and is facing towards the right. Current neighbour positions are shown in blue and current velocities are shown in green. The ground-truth choice is highlighted in light green. (a) In the first row, the decision of the network is strongly influenced by the collision-avoidance and occupancy map of the DCM.  Consequently, the pedestrian changes the direction of motion and turns left maintaining constant speed. (b) In the second row, the leader-follower map exerts a strong influence on the final decision-making causing the model to choose the anchor corresponding to slowing-down. (c) In the third row, the leader-follower map is not strong in intensity and the neural network map guides the decision making resulting in the model maintaining constant speed.}
\label{fig:viz}
\end{figure*}

We demonstrate the ability of our network to output interpretable intents in Fig.~\ref{fig:viz}. The direction of the pedestrian of interest is normalized and is facing towards the right. For each row in Fig.~\ref{fig:viz}, in addition to the ground-truth map (leftmost), we illustrate the activation maps of: all combined factors, the neural network (NN) map, the overall DCM map and finally the dominant behavioral rules that comprise the DCM function, according to the presented scene. In the first row, we observe that the model correctly chooses to turn left while maintaining constant speed. The different activation maps help to explain the rationale behind the model's decision. Indeed, due to the increased number of potentially colliding neighbours, one can observe that the collision avoidance map along with the occupancy map exerts a strong influence on the decision-making, resulting in the network outputting desired choice of intent. 

In the second and third row, we demonstrate two similar cases of leader-follower (LF) that results in different network outputs. In the former case, one of the neighbours being close to the pedestrian of interest results in the LF map exhibiting a strong affinity for slowing down. The strength of the LF map is strong enough to overturn the NN map's decision to maintain constant speed. In contrast, in the third row, the influence of the LF map is weaker. Therefore, due to the preference of NN map, the overall network chooses to maintain constant speed and direction. Thus, we observe that the DCM maps work well in conjunction with the NN map to provide interpretable outputs.




\newpage



\subsection{Direction Normalization}\label{sec:dir_norm}
\vspace{-0.5em}
Direction normalization at every time-step is an necessary step to enable the integration of the DCM framework. According to the DCM framework for pedestrian forecasting \cite{ROBIN200936}, the anchor set $A$ at each time-step is defined dynamically with respect to the current speed and direction of motion. The input scene needs to be rotated so that the pedestrian of interest faces the same direction at every time-step and consequently the appropriate anchor can be chosen by the model. Therefore, thanks to this normalization, we can successfully incorporate the interpretability of DCM without compromising prediction accuracy.

We argue that direction normalization is a general normalization scheme that provides a performance boost, in terms of avoiding collisions, when applied to many existing trajectory forecasting models. The reason behind the improvement is that direction normalization makes the forecasting model rotation-invariant at each time-step, thus allowing the model to focus explicitly on learning the social interactions. We would like to note that the direction normalization differs from the one proposed in \cite{Chai2019MultiPathMP}, as we rotate the direction of motion of a pedestrian's model at each time-step and not just at the end of observation. 
\vspace{1em}

To verify the efficacy of direction normalization, we perform a comparison between various baselines and their direction-normalized versions. Table~\ref{Norm_Synth} and Table~\ref{Norm_Real} illustrate the performance boost obtained on applying direction normalization to different trajectory prediction models on both TrajNet++ synthetic and real dataset. On the synthetic dataset, we observe that our proposed normalization scheme provides performance improvement on all the metrics across all the models. On the real dataset, we observe that direction normalization improves the model prediction collision performance. 

\begin{figure}[hbt!]
\centering
\begin{subfigure}[hbt!]{0.45\textwidth}
    \includegraphics[width=\textwidth]{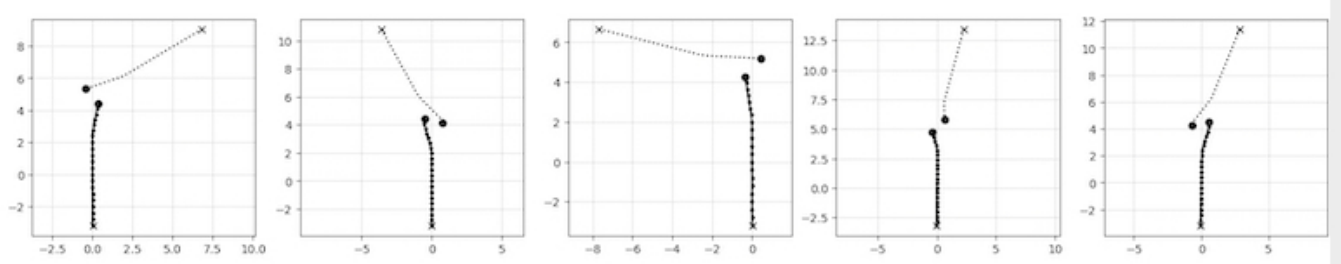}
    \includegraphics[width=\textwidth]{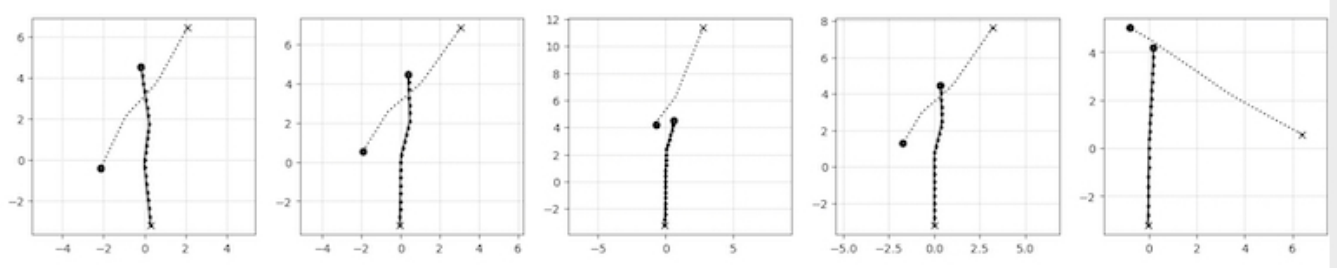}
    \caption{Nearest Neighbours of baseline in representation space}
\end{subfigure}
\begin{subfigure}[hbt!]{0.45\textwidth}
    \includegraphics[width=\textwidth]{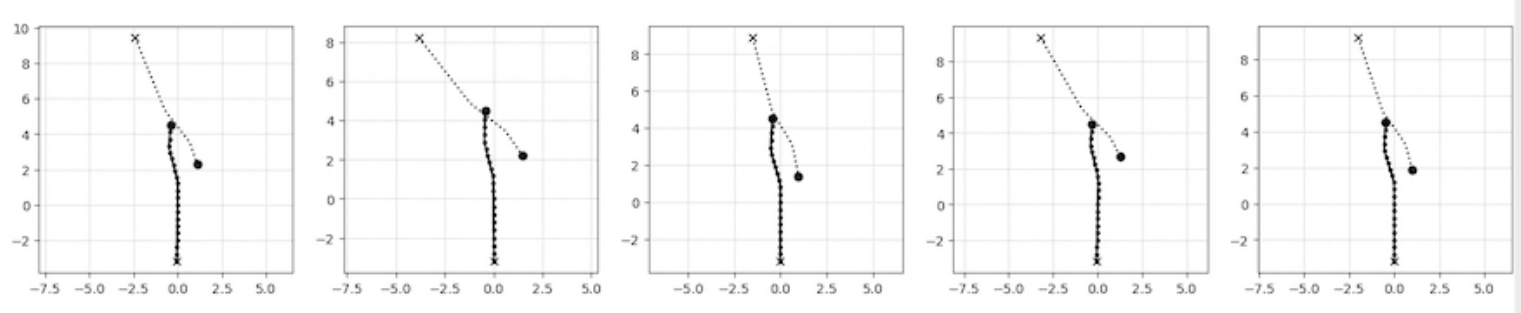}
    \includegraphics[width=\textwidth]{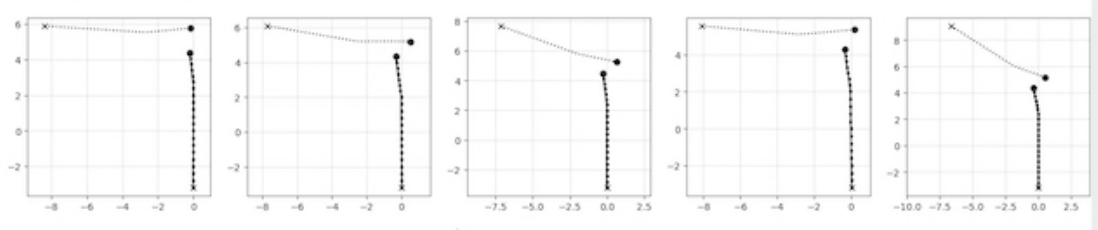}
    \caption{Nearest Neighbour of direction normalized baseline}
\end{subfigure}
\caption{Semantically similar representations are obtained on training networks using direction normalization.}
\label{fig:semantics}
\end{figure}

In addition to providing a performance improvement, the latent representations obtained by a network trained using direction normalization are semantically meaningful in the aspect of modelling social interactions. To demonstrate this, we consider a toy dataset of two pedestrians interacting with each other. The two pedestrians are initialized at different positions on the circumference of a circle with the objective of reaching the diametrically opposite position. The two pedestrians interact at different angles and positions. We train a S-LSTM \cite{Alahi2016SocialLH} and direction-normalized S-LSTM model on this dataset. During testing, we obtain the representation outputted by the LSTM encoder for the particular testing scene and find the closest latent-space representations in the training set. Fig.~\ref{fig:semantics} represents the top-4 nearest neighbours, in the latent-space, from the training set. We observe that the direction-normalized representations are more semantically similar in terms of not only the trajectory of the pedestrian of interest but also the neighbourhood configuration around the pedestrian. 

\begin{table}[h]
\centering
\resizebox{0.45\textwidth}{!}{
\begin{tabular}{ |c|c|c|c| } 
 \hline
\textbf{Model} & \textbf{ADE / FDE} & \textbf{Col-I} & \textbf{Top-3 ADE / FDE}\\ 
 \hline
 \hline
 \multicolumn{4}{|c|}{\textbf{Unimodal methods}}\\
\hline
 NN-LSTM \cite{Kothari2020HumanTF} & 0.25/0.50 & 1.24 & 0.25/0.50 \\
 \hline
 NN-LSTM (N) & \textbf{0.20/0.43} & \textbf{0.1} & \textbf{0.20/0.43} \\
 \hline
  \hline
  \multicolumn{4}{|c|}{\textbf{Multimodal methods}}\\
 \hline
 WTA \cite{Rupprecht2016LearningIA} & 0.28/0.54 & 4.8	& 0.22/0.42 \\
 \hline
 WTA (N) &\textbf{0.22/0.45} & \textbf{0.6} & \textbf{0.17/0.35} \\
 \hline
 SGAN \cite{Gupta2018SocialGS} & 0.27/0.54 & 5.1 & 0.22/0.43 \\
 \hline
 SGAN (N) & \textbf{0.24/0.50} & \textbf{1.4} & \textbf{0.19/0.37} \\
 \hline
  CVAE \cite{Lee2017DESIREDF} & 0.26/0.52 & 1.9 & 0.23/0.47 \\
 \hline
  CVAE (N) & \textbf{0.23/0.47} & \textbf{0.5} & \textbf{0.22/0.45} \\
 \hline
\end{tabular} }
\setlength{\belowcaptionskip}{-6pt}
\caption{Effect of normalization on synthetic data. Errors reported are ADE / FDE in meters, Col I in \%. (N) represents direction-normalized version of the baseline.}
\label{Norm_Synth}
\end{table}

\begin{table}[h]
\centering
\resizebox{0.45\textwidth}{!}{
\begin{tabular}{ |c|c|c|c| } 
 \hline
\textbf{Model} & \textbf{ADE / FDE} & \textbf{Col-I} & \textbf{Top-3 ADE / FDE}\\ 
 \hline
 \hline
 \multicolumn{4}{|c|}{\textbf{Unimodal methods}}\\
\hline
 NN-LSTM \cite{Kothari2020HumanTF} & 0.58/1.24 & 7.5 (0.25) & 0.58/1.24* \\
 \hline
 NN-LSTM (N) & 0.63/1.36 & \textbf{5.9} & 0.63/1.36* \\
 \hline
 D-LSTM \cite{Kothari2020HumanTF} & 0.57/1.24 & 5.5 (0.19) & 0.57/1.24 \\
 \hline
 D-LSTM (N)  & 0.62/1.32 & \textbf{4.5} & 0.62/1.32 \\
 \hline
  \hline
  \multicolumn{4}{|c|}{\textbf{Multimodal methods}}\\
 \hline
 WTA \cite{Rupprecht2016LearningIA} & 0.65/1.46 & 5.1 & \textbf{0.49/1.05} \\
 \hline
 WTA (N)& 0.63/1.38 & \textbf{4.4} & 0.54/1.15 \\
\hline
 SGAN \cite{Gupta2018SocialGS} & 0.66/1.45 & 5.9 & 0.51/1.08 \\
 \hline
 SGAN (N) & 0.64/1.38 & \textbf{4.0} & 0.51/1.07 \\
 \hline
 CVAE \cite{Lee2017DESIREDF} & 0.60/1.28 & 5.7 & 0.55/1.20 \\
 \hline
 CVAE (N) & 0.62/1.34 & \textbf{4.2} & 0.58/1.23 \\
\hline
\end{tabular}}
\setlength{\belowcaptionskip}{-10pt}
\caption{Effect of normalization on real data. Errors reported are ADE / FDE in meters, Col I in \%. (N) represents direction-normalized version of the baseline.}
\label{Norm_Real}
\end{table}

\newpage

\section{Conclusions}
We approach the task of human trajectory forecasting by disentangling human motion into high-level discrete intents and low-level scene-specific refinements. By leveraging recent works in hybrid choice models, the discretized intents are selected using both interpretable knowledge-based functions and neural network predictions from the scene. While the former allows us to understand which human motion rules are present in predicting the next intent, the latter handles the effects of both long term dependencies and complex human interactions. Through experiments on both synthetic and real data, we highlight not only the interpretability of our method, but also the accurate predictions outputted by our model. This is made possible because of the scene-specific refinements which efficiently cast the discrete problem into the continuous domain. 


\section{Acknowledgements}

This work was supported by the Swiss National Science Foundation under the Grant 2OOO21-L92326, EPFL Open Science fund, and Honda R\&D Co., Ltd. We also thank VITA members and reviewers for their valuable comments. 

\FloatBarrier
{\small
\bibliographystyle{ieee_fullname}
\bibliography{egbib}
}

\end{document}